\documentclass{article}[11pt]
\usepackage{graphicx} 

\usepackage{amsmath}
\usepackage{amssymb} 

\usepackage{authblk} 

\usepackage[utf8]{inputenc}
\usepackage{graphicx}
\usepackage{enumitem}
\usepackage{natbib}
\usepackage{amsmath,amsfonts,amssymb,amsthm}
\usepackage{mathtools}
\usepackage[margin=3cm]{geometry}
\usepackage{algorithm}
\usepackage{algpseudocode}
\usepackage{hyperref}
\usepackage{cleveref}
\usepackage{autonum}
\usepackage{xcolor}
\usepackage{upgreek}

\usepackage{hyperref}
\usepackage{xcolor}

\hypersetup{
     colorlinks=true,
     citecolor=blue,  
    linkbordercolor=black, 
}

\theoremstyle{plain}
\newtheorem{theorem}{Theorem}[section]
\newtheorem{proposition}[theorem]{Proposition}

\newtheorem{corollary}[theorem]{Corollary}
\theoremstyle{definition}

\theoremstyle{remark}

\newcommand{\rset}{\mathbb{R}}

\newcommand{\rmd}{\mathrm{d}}

\newcommand{\Id}{\mathrm{Id}}

\newcommand{\rmC}{\mathrm{C}}

\newcommand{\sigmadata}{\sigma_{\text{tar}}}

\def \TSI {\mathrm{TSI}}
\def \TSM {\mathrm{TSM}}
\def \DSI {\mathrm{DSI}}
\def \DSM {\mathrm{DSM}}
\def \Var {\mathrm{Var}}

\title{Target Score Matching}
\author[1]{Valentin De Bortoli\thanks{vdebortoli@google.com,~mhutchin@google.com,~pewi@google.com,~arnauddoucet@google.com}} 
\author[2]{Michael Hutchinson}
\author[2]{Peter Wirnsberger}
\author[1]{Arnaud Doucet} 

\affil[1]{Google DeepMind}
\affil[2]{Isomorphic Labs}

\date{}

\begin{document}
\maketitle

\begin{abstract}
Denoising Score Matching estimates the score of a ``noised'' version of a target distribution by minimizing a regression loss and is widely used to train the popular class of Denoising Diffusion Models. A well known limitation of Denoising Score Matching, however, is that it yields poor estimates of the score at low noise levels. This issue is particularly unfavourable for problems in the physical sciences and for Monte Carlo sampling tasks for which the score of the ``clean'' original target is known. Intuitively, estimating the score of a slightly noised version of the target should be a simple task in such cases. In this paper, we address this shortcoming and show that it is indeed possible to leverage knowledge of the target score. We present a Target Score Identity and corresponding Target Score Matching regression loss which allows us to obtain score estimates admitting favourable properties at low noise levels.
\end{abstract}


\section{Introduction and Motivation}
\subsection{Denoising Score Identity and Denoising Score Matching}
Consider a $\mathbb{R}^d$-valued random variable $X \sim p_X$ and let $Y|(X=x) \sim p_{Y|X}(\cdot|x)$ be a ``noisy'' version of $X$. We denote the joint density of $(X,Y)$ by $p_{X,Y}(x,y)=p_X(x) p_{Y|X}(y|x)$ and refer to expectation and variance w.r.t. to this distribution as $\mathbb{E}_{X,Y}$ and $\textup{Var}_{X,Y}$, respectively. We are interested in estimating the score of the distribution of $Y$, that is $\nabla \log p_Y(y)$, where  
\begin{equation}\label{eq:marginalY}
    p_Y(y)= \int p_X(x)p_{Y|X}(y|x) \rmd x.
\end{equation}
Evaluating this score is particularly useful for denoising tasks, especially Denoising Diffusion Models (DDM) which require estimates at different noise levels~\citep{sohl2015deep,ho2020denoising,song2020score}. 
A standard derivation shows that the identity
\begin{equation}\label{eq:DSMidentity}
    \nabla \log p_Y(y) =\int \nabla \log p_{Y|X}(y|x)~p_{X|Y}(x|y) \rmd x
\end{equation}
holds under mild regularity assumptions, where henceforth $\nabla \log p_{Y|X}(y|x)$ denotes the gradient with respect to $y$ and 
\begin{equation}
    p_{X|Y}(x|y)=\frac{p_X(x)p_{Y|X}(y|x)}{p_Y(y)}
\end{equation}
is the posterior density of $X$ given $Y=y$ (see, for example, \cite{vincent2011connection}). We will refer to \eqref{eq:DSMidentity} as the \emph{Denoising Score Identity} ($\DSI$). In scenarios where it is possible to compute $\nabla \log p_{Y|X}(y|x)$ and $p_{X|Y}(x|y)$ is known up to a normalizing constant, the score can then be approximated by estimating the expectation in \eqref{eq:DSMidentity} using Markov chain Monte Carlo (MCMC); see e.g. Appendix D.3 in \citep{vargasDDSampler2023} and \citep{huang2023monte}.

However, in most standard generative modeling applications, we only have access to $\nabla \log p_{Y|X}(y|x)$ and i.i.d. samples $(X_i,Y_i)_{i=1}^n$ from $p_{X,Y}(x,y)$ and do not know $p_{X|Y}(x|y)$. In this context, \emph{Denoising Score Matching} ($\DSM$) \citep{vincent2011connection} leverages $\DSI$ \eqref{eq:DSMidentity} by approximating $\nabla \log p_Y(y)$ using some function $s_Y^\theta(y)$ whose parameters are obtained by minimizing the regression loss
\begin{equation}\label{eq:DSMloss}
    \ell_{\textup{DSM}}(\theta)=\mathbb{E}_{X,Y}[||s_Y^\theta(Y) - \nabla \log p_{Y|X}(Y|X)||^2],
\end{equation}
which is in practice approximated using samples $(X_i,Y_i)_{i=1}^n$. $\DSM$ is an alternative to Implicit Score Matching \citep{Hyvarinen:2005a} which only requires access to noisy samples $(Y_i)_{i=1}^n$ from $p_Y(y)$ but is computationally more expensive in high dimensions as optimizing the corresponding loss requires computing the gradient of the divergence of a $d$-dimensional vector field.




\subsection{Limitations}
\label{sec:limitations}

Consider the case where $Y$ is obtained by adding some independent noise $W$ to $X$, i.e.
\begin{equation} \label{eq:additive}
Y=X+ W,\qquad W \sim p_W(\cdot).
\end{equation}
If one can sample from $p_{X|Y}(x|y)$, $\DSI$ \eqref{eq:DSMidentity} suggests a Monte Carlo estimator of the score $\nabla \log p_Y(y)$ obtained by averaging $\nabla \log p_{Y|X}(y|X)$ over samples $X \sim p_{X|Y}(\cdot|y)$. In the case of Gaussian noise, $p_W(w)=\mathcal{N}(w;0,\sigma^2 I)$, we have that $\sum_{i=1}^d \Var_{X|Y}((\nabla \log p_{Y|X}(Y|X))_i) \sim d \sigma^{-2}$ as $\sigma \to 0$. So the variance of such an estimator is higher for low noise levels.

The high variance of the Monte Carlo estimator for low noise levels is an independent issue to the high variance of the $\DSM$ regression loss used to approximate this estimator. Indeed, while the original $\DSM$ loss also exhibits high variance at low noise levels, we can re-arrange \eqref{eq:DSMidentity} to obtain the so-called \emph{Tweedie identity}  \citep{robbins1956empirical,miyasawa1961empirical,raphan2011least}
\begin{equation}\label{eq:Tweedie}
\mathbb{E}[X|Y=y]=y+\sigma^2 \nabla \log p_Y(y).
\end{equation}
This identity provides us with an alternative way for computing the score by estimating $ \mathbb{E}[X|Y=y]$, using a regression loss of the form $\mathbb{E}_{X,Y}[||x^{\theta}(Y)-X||^2]$. In this case, the regression target is simply $X$ and therefore does not exhibit exploding variance as $\sigma \to 0$. However, this approximation $x^\theta$ is used to compute the score as \begin{equation}
    s^\theta(y) = (x^\theta(y) - y) / \sigma^2 . 
\end{equation} 
Hence the error of $x^\theta$ is amplified as $\sigma \to 0$. In fact, this approach is simply equivalent to a rescaling of the $\DSM$ loss as $\mathbb{E}_{X,Y}[||x^{\theta}(Y)-X||^2]=\sigma^4 \ell_{\textup{DSM}}(\theta)$. In the DDM literature, this reparameterisation is sometimes called the $x_0$-prediction. Many other regression targets have been proposed in the context of diffusion models, see \cite{salimans2022progressive} for instance. All these parameterisations exhibit the same issues as the direct score prediction or the  $x_0$-prediction, since the resulting score approximation $s_\theta$ exhibits exploding variance as $\sigma \to 0$. 

Other techniques have also been proposed in order to derive better behaved regression losses, see for example \cite{wang2020wasserstein} and \cite{karras2022elucidating}. While these works mitigate the high variance of the regression target, we emphasize that they fail to address the fundamental variance issue of the score estimator itself.

\section{Target Score Identity and Target Score Matching}
\label{sec:target_informed_score_matching}

We will focus hereafter on scenarios where the score $\nabla \log p_X(x)$ of the ``clean'' target $p_X(x)$ can be computed exactly. As mentioned earlier, this is not the case for most generative modeling applications where $p_X(x)$ is only available through samples. However, the score $\nabla \log p_X(x)$ is known in many physical science applications and Monte Carlo sampling tasks that are actively investigated using denoising techniques; see e.g.  \citet{zhang2021path,arts2023two,cotler2023renormalizing,herron2023inferring,vargasDDSampler2023,vargas2023bayesian,wang2023generative,zhang2023diffusion,zheng2023towards,akhoundsadegh2024iterated,huang2023monte,phillips2024,richter2023improved}.

For ease of presentation, we restrict ourselves to the additive noise model \eqref{eq:additive} in this section and discuss a more general setup in the next section. At low noise levels, we expect $\nabla \log p_Y(y) \approx \nabla \log p_X(y)$ yet neither the $\DSI$ \eqref{eq:DSMidentity} nor the $\DSM$ loss \eqref{eq:DSMloss} take advantage of this fact. On the contrary, the  \emph{Target Score Identity} ($\TSI$) and the \emph{Target Score Matching} ($\TSI$) loss that we present below address this shortcoming and leverage explicit knowledge of $\nabla \log p_X(x)$. Henceforth, we assume that all the regularity conditions allowing us to differentiate the densities and interchange differentiation and integration are satisfied. 
\begin{proposition}
\label{prop:ti_identity}
For the additive noise model \eqref{eq:additive}, the following \emph{Target Score Identity} holds
\begin{equation}\label{eq:TIscoreidentity}
\nabla \log p_Y(y)= \int \nabla \log p_X(x)~ p_{X|Y}(x|y) \rmd x.
\end{equation}
\end{proposition}
\begin{corollary}
By symmetry, we also have 
\begin{equation}
\nabla \log p_Y(y)= \int \nabla \log p_W(w)~ p_{W|Y}(w|y) \rmd w=\int \nabla \log p_W(y-x)~ p_{X|Y}(x|y) \rmd x,
\end{equation}
which is $\DSI$ for \eqref{eq:additive}.

\end{corollary}
The proof of this result and all other results in the main paper are given in Appendix \ref{sec:Mainproofs}. An alternative proof of this identity for additive Gaussian noise relying on the Fokker--Planck equation is provided in Appendix \ref{sec:FP_proof_ti}.

This result is not new, which is to be expected given its simplicity. It is part of the folklore in information theory and can be found, for example, in \citep{blachman1965convolution}\footnote{In machine learning, it was used (and derived independently) in Appendix C.1.3. of \citep{debortoli2021diffusion} to establish some theoretical properties of DDM.}.
However, to the recent exception of \cite{phillips2024}, the remarkable computational implications of this identity do not appear to have been exploited previously. As discussed further, \cite{akhoundsadegh2024iterated} also rely implicitly on this identity. $\TSI$ shows that, if $p_{X|Y}(x|y)$ is known pointwise up to a normalizing constant, then the score can be estimated by using an Importance Sampling (IS) or MCMC approximation of $p_{X|Y}(x|y)$ to compute the expectation \eqref{eq:TIscoreidentity}. The integrand $\nabla \log p_X(x)$ will typically have much smaller variance under $p_{X|Y}$ than the integrand $\nabla \log p_{Y|X}(y|x)$ appearing in \eqref{eq:DSMidentity} at low noise levels.

Having access to samples $(X_i,Y_i)_{i=1}^n$ from $p_{X,Y}$, we can also can estimate the score $\nabla \log p_Y(y)$ by minimizing the following \emph{Target Score Matching} ($\TSM$) regression loss.
\begin{proposition}
\label{prop:ti_loss}
Consider a class of regression functions $s^\theta_Y:\mathbb{R}^d \rightarrow \mathbb{R}^d$ for $\theta \in \Theta$. For the additive noise model \eqref{eq:additive}, we can estimate the score $\nabla \log p_Y(y)$ by minimizing the following regression loss
\begin{equation}\label{eq:TIregressionloss}
    \ell_{\textup{TSM}}(\theta)=\mathbb{E}_{X,Y}[||s_Y^\theta(Y) - \nabla \log p_{X}(X)||^2].
\end{equation}
Additionally, the $\TSM$ loss $\ell_{\textup{TSM}}(\theta)$ and the $\DSM$ loss $\ell_{\textup{DSM}}(\theta)$ satisfy
\begin{equation}\label{eq:TIandDSM}
    \ell_{\textup{TSM}}(\theta)=\ell_{\textup{DSM}}(\theta)+ \int \|\nabla \log p_X(x) \|^2 p_X(x) \rmd x -\int \| \nabla \log p_{Y|X}(y|x) \|^2 p_{X,Y}(x,y) \rmd x \rmd y.
\end{equation}
\end{proposition}
Contrary to $\ell_{\textup{DSM}}(\theta)$, $\ell_{\textup{TSM}}(\theta)$ does not require having access to $\nabla \log p_{Y|X}(y|x)=\nabla \log p_W(y-x)$. This can be useful when the score of the noise distribution is not analytically tractable; see e.g. Section \ref{sec:extensions} for applications to Riemmanian manifolds. Finally, we note that the relationship \eqref{eq:TIandDSM} shows that $\ell_{\textup{DSM}}(\theta)$ takes large values compared to $\ell_{\textup{TSM}}(\theta)$ at low noise levels as both quantities are positive and the expected conditional Fisher information, $\mathbb{E}_{X,Y}[ \| \nabla \log p_{Y|X}(Y|X) \|^2]$, takes large values.

\begin{corollary}
In many applications, the observation model of interest is of the form 
\begin{equation}\label{eq:additivescale}
    Y=\alpha X + W
\end{equation}
for $\alpha \neq 0$ and $W$ independent of $X$. In this case, $\TSI$ becomes
\begin{equation}\label{eq:TIscoreidentityscale}
\nabla \log p_Y(y)=\alpha^{-1} \int \nabla \log p_X(x)~ p_{X|Y}(x|y) \rmd x,
\end{equation}
while $\TSM$ is given by 
\begin{equation}\label{eq:TIregressionlossscale}
    \ell_{\textup{TSM}}(\theta)=\mathbb{E}_{X,Y}[||s_Y^\theta(Y) -\alpha^{-1} \nabla \log p_{X}(X)||^2].
\end{equation}
\end{corollary}

For model \eqref{eq:additivescale}, if the score is estimated through $\TSM$, it is thus sensible to use a parameterization for $s^\theta_Y$ of the form $s^\theta_Y(y)=\alpha^{-1} \nabla \log p_X(y)+\epsilon^\theta(y)$.

In practice, we can consider any convex combination of $\DSI$ (equation \eqref{eq:DSMidentity}) and $\TSI$ (equation \eqref{eq:TIscoreidentityscale}) to obtain another score identity which we can use to derive a score matching loss\footnote{We can alternatively consider a convex combination of the $\DSM$ and $\TSM$ losses.}. The score identity considered by \cite{phillips2024} and corresponding score matching loss follow this approach. Therein one considers a variance preserving denoising diffusion model \citep{song2020score} where 
$Y_t=\alpha_t X + \sqrt{1-\alpha^2_t} \epsilon$
for $\epsilon \sim \mathcal{N}(0,I)$ and $(\alpha_t)_{t\in [0,1]}$ a continuous decreasing function such that $\alpha_0=1$ and $\alpha_1 \approx 0$. In this case, $\alpha^{-1}_t \nabla \log p_X(x)$ will typically be better behaved than $\nabla \log p_{Y_t|X}(y|x)$ for $t$ close to $0$ and vice-versa for $t$ close to $1$. \cite{phillips2024} exploits this behaviour to propose the score identity 
\begin{equation}\label{eq:scorePhillips}
    \nabla \log p_{Y_t}(y)=\int [\alpha_t (x+ \nabla \log p_X(x))-y]~ p_{X|Y_t}(x|y)\rmd x,
\end{equation}
which is the sum of $\alpha^2_t$ times $\TSI$ (equation \eqref{eq:TIscoreidentityscale}) and $1-\alpha^2_t$ times $\DSI$ (equation \eqref{eq:DSMidentity}). They use the integrand in \eqref{eq:scorePhillips} to define a score matching loss. The rationale for this choice is that the integrand will be close to the true score for $t\approx 0$ and $t \approx 1$ when $X \sim p_{X|Y_t}$. In practice, the ``best'' loss function one can consider will be a function of the target $p_X$, ``noise'' $p_{Y|X}$ and $\alpha$. In Appendix \ref{sec:preconditioning_training}, we follow the analysis \cite{karras2022elucidating} to derive a loss admitting desirable properties in a simplified Gaussian setting. 

We finally note that the very recent score estimate proposed in \citep{akhoundsadegh2024iterated} (see Eq. (8) therein) can be reinterpreted as a self-normalized IS approximation in disguise of $\TSI$. It considers for the model \eqref{eq:additive} and $W \sim \mathcal{N}(0,\sigma^2 I)$ the IS proposal distribution $q(x)=\mathcal{N}(x;y,\sigma^2 I)=p_W(y-x)$ to approximate $p_{X|Y}(x|y)\propto p_X(x)p_W(y-x)$. For $\sigma \ll 1$, this importance distribution will perform well as $p_{X|Y}(x|y)$ is concentrated around $y$. For larger $\sigma$, the variance of the resulting IS estimate could be significant.

\section{Extensions}\label{sec:extensions}

Next, we present a few extensions of $\TSI$ and $\TSM$.

\subsection{Extension to non-Additive Noise}
We consider here a general noising process defined by 
\begin{equation}\label{eq:generalnoise}
    p_{Y|X}(y | x) = F(\Phi(y, x)),
\end{equation}
where we assume that $\Phi(y, \cdot)$ is a $\rmC^2$-diffeomorphism for any $y \in \rset^d$ and that $\Phi$ is smooth. We denote by $\nabla_1$, respectively $\nabla_2$, the gradient of $\Phi$ or $\Phi^{-1}$ with respect to its first, respectively second, argument.

\begin{proposition}\label{prop:generalnoise} For the noise model \eqref{eq:generalnoise}, the following \emph{Target Score Identity} holds
\begin{align}
    \nabla \log p_Y(y) = \int [\nabla_1 \Phi^{-1}(y, \Phi(y, x))^\top \nabla \log p_X(x)  + \nabla_y \log |\det(\nabla_2 \Phi^{-1}(y, \cdot))| (\Phi^{-1}(y, x)) ]  p_{X|Y}(x| y) \rmd x  .
\end{align}
\end{proposition}
For $\Phi(y,x)=y-\alpha x$ and $F(w)= p_W(w)$, we have $\Phi^{-1}(y,z) = (y-z)/\alpha$ and therefore one has $\nabla_1 \Phi^{-1}(y, \Phi(y, x))^\top = \Id / \alpha$ and $\nabla_y \log |\det(\nabla_2 \Phi^{-1}(y, \cdot))| = 0$. Hence, we recover \eqref{eq:TIscoreidentityscale}. 
We can thus estimate the score by minimizing the following $\TSM$ loss \begin{equation}\label{eq:generalTSMloss}
    \ell_{\textup{TSM}}(\theta)=\mathbb{E}_{X,Y}[||s^\theta_Y(Y)-[\nabla_1 \Phi^{-1}(Y, \Phi(Y, X))^\top \nabla \log p_X(X)  + \nabla_y \log |\det(\nabla_2 \Phi^{-1}(Y, \cdot))| (\Phi^{-1}(Y, X)) ]||^2].
\end{equation}

\subsection{Extension to Lie groups}
Consider a Lie group $G$ which admits a bi-invariant metric. We denote by $\mu$ the (left) Haar measure on $G$. Let $p_X$ denotes the density of $X$ w.r.t. $\mu$. We assume the following additive model on the Lie group, i.e.~ $Y = X +_G W$, where $+_G$ is the group addition on $G$ and $W \sim p_W$ with density $p_W$ w.r.t. $\mu$. If $G = \rset^d$, we recover the Euclidean additive model of Section \ref{sec:target_informed_score_matching}. 
For any smooth $f: \ G \to G$ and $x \in G$, we denote $\mathrm{d}f(x): \ \mathrm{T}_x G \to \mathrm{T}_{f(x)} G$, the differential operator of $f$ evaluated at $x$. Similarly, for any smooth $f: \ G \to \rset$, we denote $\nabla f(x)$ its \emph{Riemannian} gradient. 

\begin{proposition}
\label{prop:ti_estimator_lie}
For a Lie group, the following \emph{Target Score Identity} holds
\begin{equation}\label{eq:TIscoreidentity_lie}
\nabla \log p_Y(y)=\int \rmd R_{x^{-1}y}(x) \nabla \log p_X(x)~ p_{X|Y}(x|y) \rmd \mu(x) , 
\end{equation}
where $R_x(y) = yx$ for any $x, y \in G$. In particular, if $G$ is a matrix Lie group, we have 
\begin{equation}\label{eq:TIscoreidentity_matrix_lie}
\nabla \log p_Y(y)=\int \nabla \log p_X(x)x^{-1}y~ p_{X|Y}(x|y) \rmd \mu(x) .
\end{equation}
\end{proposition}

$\DSI$ and $\DSM$ have been extended to Riemannian manifolds in \citep{debortoli2022riemannian,huang2022riemannian}; see e.g. \cite{watson2023novo} for an application to protein modeling. Leveraging the Lie group structure to obtain a tractable expression of the heat kernel defining $p_{Y|X}$ and therefore obtain more amenable $\DSI$ and $\DSM$ was considered in \citep{yim2023se,lou2023scaling,leach2022denoising}. Contrary to these works, we do not need to know the exact form of the additive noising process $p_W$. 

Adapting $\DSI$ and $\DSM$ to Riemmanian manifolds simply requires replacing the Euclidean gradient by the Riemannian gradient. This is not the case for $\TSI$, i.e. Proposition \ref{prop:ti_estimator_lie} is not obtained by replacing the Euclidean gradient by the Riemannian gradient in Proposition \ref{prop:ti_identity}. This is because, while $s_Y^\theta(y) \in \mathrm{T}_yG$, where $\mathrm{T}_yG$ is the tangent space of $G$ at $y$. However, we have that $\nabla \log p_X(x)  \in \mathrm{T}_xG$ and therefore these two quantities are not immediately comparable and we use $\rmd R_{x^{-1}y}(x)$ which transports $\mathrm{T}_x G$ onto $\mathrm{T}_y G$. In contrast, in the case of $\DSI$, both $s_Y^\theta(y) \in \mathrm{T}_yG$ and $\nabla_y \log p_{Y|X}(y|x) \in \mathrm{T}_yG$. It is also possible to extend straightforwardly Proposition \ref{prop:ti_loss} to the context of Lie groups.

\begin{proposition}
\label{prop:ti_loss_lie}
Consider a class of regression functions such that for any $y \in G$, $s_\theta(y) \in \mathrm{T}_yG$ for $\theta \in \Theta$. We can estimate the score $\nabla \log p_Y(y)$ by minimizing the $\TSM$ regression loss
\begin{equation}\label{eq:TIregressionloss_lie}
    \ell_{\textup{TSM}}(\theta)=\mathbb{E}_{X,Y}[||s_Y^\theta(Y) - \rmd R_{X^{-1}Y}(X) \nabla \log p_{X}(X)||^2] , 
\end{equation}
where $R_x(y) = yx$ for any $x, y \in G$.

\end{proposition}

\subsection{Extension to Bridge Matching}
Let $Y$ be given by
\begin{equation}\label{eq:interpolant}
    Y=\alpha X_0 + (1-\alpha) X_1 + W,
\end{equation}
where $X_0 \sim p_{X_0}$, $X_1 \sim p_{X_1}$, $W \sim p_W$ are independent and $0<\alpha<1$. We are interested in evaluating the score $\nabla \log p_Y(y)$. In the context of generative modeling, \eqref{eq:interpolant} appears when one builds a transport map bridging $p_{X_0}$ to $p_{X_1}$; see e.g. \citep{peluchettinon2021,liu2022let,albergo2023stochastic,lipman2022flow}. We are again considering here a scenario where we have access to the exact scores of $p_{X_0}$ and $p_{X_1}$. 

\begin{proposition}\label{prop:interpolant}
For the model \eqref{eq:interpolant}, the following \emph{Target Score Identity} holds
\begin{align}
\nabla \log p_Y(y)&=\alpha^{-1} \int \nabla \log p_{X_0}(x_0)~p_{X_0,X_1|Y}(x_0,x_1|y) \rmd x_0 \rmd x_1 \label{eq:scoreinterpolant1}\\
&=(1-\alpha)^{-1} \int \nabla \log p_{X_1}(x_1)~p_{X_0,X_1|Y}(x_0,x_1|y) \rmd x_0 \rmd x_1,\label{eq:scoreinterpolant2}
\end{align}
where $p_{X_0,X_1|Y}(x_0,x_1|y) \propto p_{X_0}(x_0)p_{X_1}(x_1)p_W(y-\alpha x_0-(1-\alpha x_1)) $ is the posterior density of $X_0,X_1$ given $Y=y$.
\end{proposition}
A convex combination of \eqref{eq:scoreinterpolant1} and \eqref{eq:scoreinterpolant2} yields the elegant identity 
\begin{equation}\label{eq:symmetricbridge}
    \nabla \log p_Y(y)=\int (\nabla \log p_{X_0}(x_0)+\nabla \log p_{X_1}(x_1)) ~p_{X_0,X_1|Y}(x_0,x_1|y) \rmd x_0 \rmd x_1.
\end{equation}
We can use these score identities to compute the score using MCMC if $p_{X_0,X_1|Y}(x_0,x_1|y)$ is available up to a normalizing constant. Alternatively, given samples from the joint distribution $p_{X_0,X_1,Y}(x_0,x_1,y)=p_{X_0}(x_0)p_{X_1}(x_1)p_{Y|X_0,X_1}(y|x_0,x_1)$, we can approximate the score by minimizing a regression loss, e.g. for \eqref{eq:symmetricbridge}
\begin{equation}\label{eq:symmetricbridge}
    \ell_{\textup{TSM}}(\theta)=\mathbb{E}_{X_0,X_1,Y}[||s^\theta_Y(Y)-(\nabla\log p_{X_0}(X_0)+\nabla \log p_{X_1}(X_1))||^2].
\end{equation}

\section{Experiments}

\subsection{Analytic estimators}
We explore experimentally the benefits of these novel score estimators by considering 1-d mixture of Gaussian targets defined by 
\begin{equation}
p_0(x_0) = \sum_{i=1}^N \pi_i \mathcal{N}(x_0;\mu_i, \sigma_i^2).
\end{equation}
Motivated by DDM, the noising process is defined by a ``noising'' diffusion 
\begin{equation}\label{eq:denoisingdiffusion}
\rmd X_t = f_t X_t \rmd t + g_t \rmd B_t ,
\end{equation}
where \(f_t = \tfrac{\rmd}{\rmd t} \log \alpha_t\) and \(g_t^2 = \tfrac{\rmd}{\rmd t}\sigma_t^2 - 2 f_t \sigma_t^2\) and $B_t$ is a standard Brownian motion.
Initialized at $X_0=x_0$, \eqref{eq:denoisingdiffusion} defines the following conditional distribution of $X_t$ given $X_0=x_0$, $p_{t|0}(x_t | x_0) = \mathcal{N}(x_t; \alpha_t x_0, \sigma_t^2 )$. 
In what follows, we focus on the cosine schedule where 
$\alpha_t = \cos((\uppi /2)t)$ and $\sigma_t = \sin((\uppi / 2)t)$.  
Consider $X_0 \sim p_0$ then the distribution of $X_t$ is given by 
\begin{equation}
p_t(x_t) = \sum_{i=1}^N \pi_i \mathcal{N}(x_t ; \mu_{i, t}, \sigma_{i,t}^2 ),\quad 
\mu_{i, t} = \alpha_t \mu_i, \quad \sigma_{i, t}^2 = \alpha_t^2\sigma_i^2 + \sigma_t^2.
\end{equation}
The posterior distribution of $X_0$ given $X_t=x_t$ is given by another mixture of Gaussians,
\begin{equation}\label{eq:posteriormixture}
p_{0|t}(x_0|x_t) = \sum_{i=1}^N \pi_{i,t} \mathcal{N}(x_0; \nu_{i,t}, \gamma_{i,t}^2),
\end{equation}
with
$\pi_{i, t} \propto \pi_i\tfrac{1}{\sqrt{2 \pi \sigma_{i,t}^2}}\exp\left(-\tfrac{(x_t - \mu_{i,t})^2}{2\sigma_{i,t}^2}\right), \nu_{i, t} = \mu_{i,t} + \tfrac{c_{i, t}^2}{\sigma_{i, t}^2}(x_t - \mu_{i, t}), \gamma_{i, t}^2 = \sigma_i^2 - \tfrac{c_{i, t}^4}{\sigma_{i, t}^2}$ and $c_{i, t}^2 = \alpha_t(\sigma_i^2 + \mu_i^2) - \mu_i\mu_{i,t}$.

For convenience, we define the following quantities
\begin{align}
    L_{\DSI}(x_0, x_t, t) &= \nabla \log p_{t|0}(x_t | x_0) , & \text{Denoising}  \\
    L_{\TSI}(x_0, x_t, t) &= \alpha^{-1}_t \nabla \log p_0(x_0) , & \text{Target} \\
    L_{w_t}(x_0, x_t, t)&= w_t L_{\DSI}(x_0, x_t, t) + (1-w_t)L_{\TSI}(x_0, x_t, t) ,  & \text{A } w_t \text{ mixture}
\end{align}
where $w_t \in [0,1]$ for any $t \in [0,1]$ is a mixture weight. 
As described before, for any of these targets denoted $L \in \{L_{\DSI}, L_{\TSI}, L_{w_t}\}$, we have 
\begin{equation}
    \nabla \log p_t(x_t) = \int L(x_0, x_t, t) p_{0|t}(x_0|x_t) \rmd x_t. \label{eq:st_est}
\end{equation}
We can then estimate \(\nabla \log p_t(x_t)\) via the Monte Carlo estimate
\begin{equation}\label{eq:scoreMonteCarlo}
    \nabla \log p_t(x_t) \approx \frac{1}{N}\sum_{i=1}^N L(X^i_0, x_t, t) \quad X^i_0 \sim p_{0|t}(\cdot|x_t) , 
\end{equation}
since \(p_{0|t}(x_0|x_t)\) given by \eqref{eq:posteriormixture} is easy to sample from. Note that in more complicated scenarios, we would have had to use MCMC or IS. In addition, we consider the score matching loss
\begin{equation}\label{eq:generalSMloss}
    \nabla \log p_t(x_t) \approx \arg\min_{s_\theta} \int_0^1 \lambda_t \mathbb{E}_{X_0, X_t}\left[||L(X_0, X_t, t) - s^\theta(X_t, t)||^2 \right] \rmd t ,
\end{equation}
where \(\lambda_t\) is a weighting function over time and this approximation holds true jointly over all \(t\). Picking \(L_{\text{DSI}}\) within \eqref{eq:generalSMloss} gives us the usual DSM loss, and picking the other identities give a series of novel score matching losses. In what follows, we explore two key quantities:
\begin{enumerate}
    \item The variance of the Monte Carlo estimates \eqref{eq:scoreMonteCarlo} of the score and
    \item the distribution of the estimated score matching loss functions around the true score, estimated using Monte Carlo.
\end{enumerate}


We investigate these quantities on a series of target distributions.
\begin{enumerate}
\item A unit Gaussian,
\item a gentle mixture of Gaussians,
\item a hard mixture of Gaussians with very separated modes, with the same mode variances,
\item a hard mixture of Gaussians with very separated modes, with the different mode variances.
\end{enumerate}
The unit Gaussian case we use to find appropriate values for the mixture weights \(w_t\).
Assuming that we have a target density \(\mathcal{N}(0, \sigma_\text{data}^2 I)\) it is possible to compute the \(w_t\) that minimises the variance of the mixture estimator. 
For this we recover
\begin{equation}
    \kappa_t := \frac{\sigma_t^2}{\sigma_t^2 + \alpha_t^2 ~\sigma_\text{data}^2},
\end{equation}
see \Cref{sec:interpolation}. In addition we can define another mixing weight
\begin{equation}
    \bar\kappa_t := \frac{\sigma_t^2}{\sigma_t^2 + \alpha_t^2 ~\sigma_\text{data mode}^2} .
\end{equation}
The difference between these two is that one used the variance of the whole target distribution, and the other used the mixture-weighted average of the variance of each mode in the mixture, \(\sigma_\text{data mode}^2 =\sum_i \pi_i \sigma_i^2 \). While \(\kappa\) is optimal for the unimodal Gaussian, we will see that \(\bar\kappa\) performs significantly better for mixtures of Gaussians.

\begin{figure}
    \centering
    \includegraphics{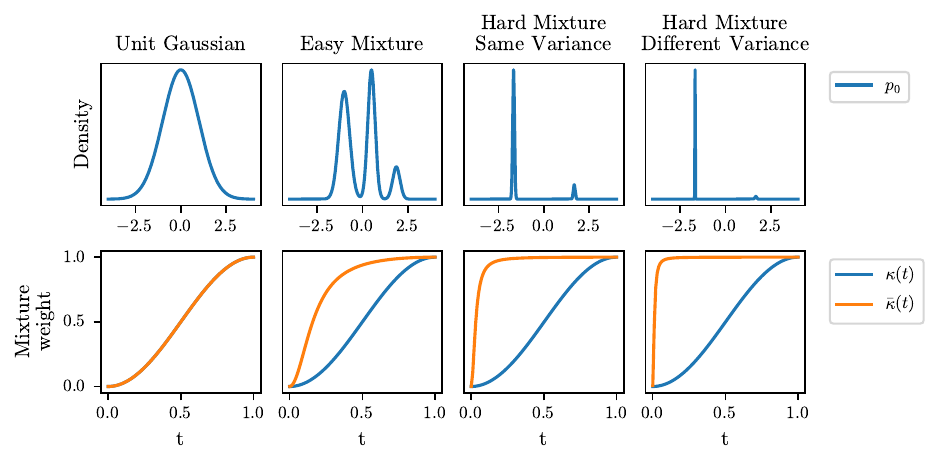}
    \caption{Target distributions (top panel), and the mixture weights \(\kappa_t\) and \(\bar\kappa_t\) through time induced by these targets (bottom panel).}
    \label{fig:p0s}
\end{figure}

\begin{figure}
    \centering
    \includegraphics{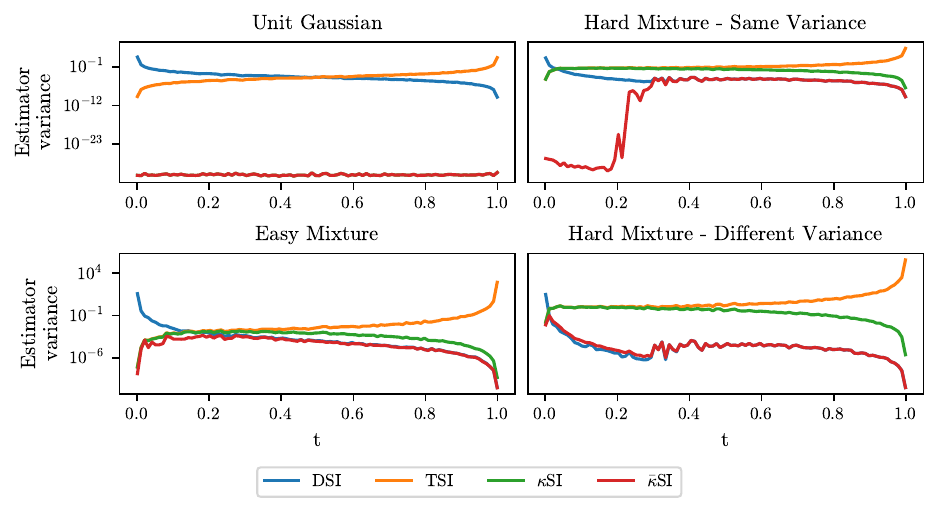}
    \caption{The estimated variance of each estimator based the score identities. Computed using 10,000 samples of the estimator. For each estimator sample, \(X_t\) is sampled from \(p_t\). For each \(X_t\), we use 100 samples from \(p_{0|t}\) to estimate the score.}
    \label{fig:si_variances}
\end{figure}

Figure \ref{fig:p0s} shows these distributions, and the mixture weightings \(\kappa_t\) and \(\bar\kappa_t\) for these distributions. 
In Figure \ref{fig:si_variances} reports the variance of the Monte Carlo estimators of the score \eqref{eq:scoreMonteCarlo} through time. For each time \(t\), we sample 10,000 samples from \(p_t\). For each of these, we estimate the score with 100 Monte Carlo samples from \(p_{0|t}\). We then compute the variance of these samples of the estimator. 

In this we see a few points of note. On the Unit Gaussian example, the Monte Carlo estimates based on $\DSI$ and $\TSI$ perform exactly oppositely, with alternatively large and small variances at \(t\approx 0\) and \(t\approx 1\). The \(\kappa_t\) and \(\bar\kappa_t\) estimators perform the same as \(\sigma_\text{data}\) and \(\sigma_{\text{data mode}}\) are the same. They also have approximately zero variance, as for this simple case the optimal mixture of the two estimators in fact gives exactly the score.

In the easy mixture case, we see the expected story. Again the $\DSI$ and $\TSI$ estimators perform opposite along time, with variance blowing up at \(t=0\) and \(t=1\) respectively. There is however a shift in when the one becomes better than the other towards \(t=0\). The \(\kappa_t\) and \(\bar\kappa_t\) estimators both perform well, with neither showing a variance blowup near \(t=0\) or \(t=1\). The \(\kappa_t\) mixture is not quite as good as the $\DSI$ estimator for \(t \geq 0.2\), but the \(\bar\kappa_t\) performs exactly as well, showing that this is the correct optimal mixture. 

In the hard mixture cases, we see again the optimal switching point between $\DSI$ and $\TSI$ move further towards \(t=0\). We again see that the \(\kappa_t\) mixture is not optimal, but that \(\bar\kappa_t\) gives us the best of both $\DSI$ and $\TSI$. Even though the time frame in which $\TSI$ is better than DSI is small, in this region we see that the $\DSI$ variance blows up and would cause difficulty estimating the score well. One thing of note is that in the hard mixture with the same mode variance the \(\bar\kappa_t\) estimator variance becomes almost zero. In this case as the modes are so far separated that as \(t\to 0\), \(p_{0|t}\) effectively becomes uni-model, bringing us back to the optimal mixture case.

\begin{figure}
    \centering
    \includegraphics{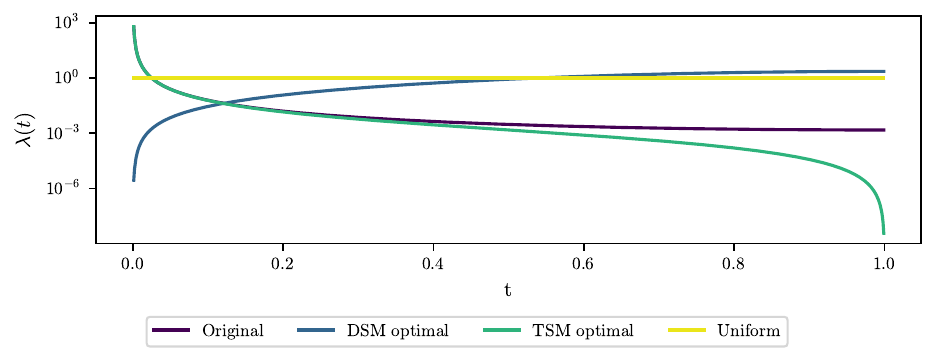}
    \caption{The different weighting functions across time, for a \(\sigma_\text{data}^2 = 1\)}
    \label{fig:weightings}
\end{figure}

\begin{figure}
    \centering
    \includegraphics{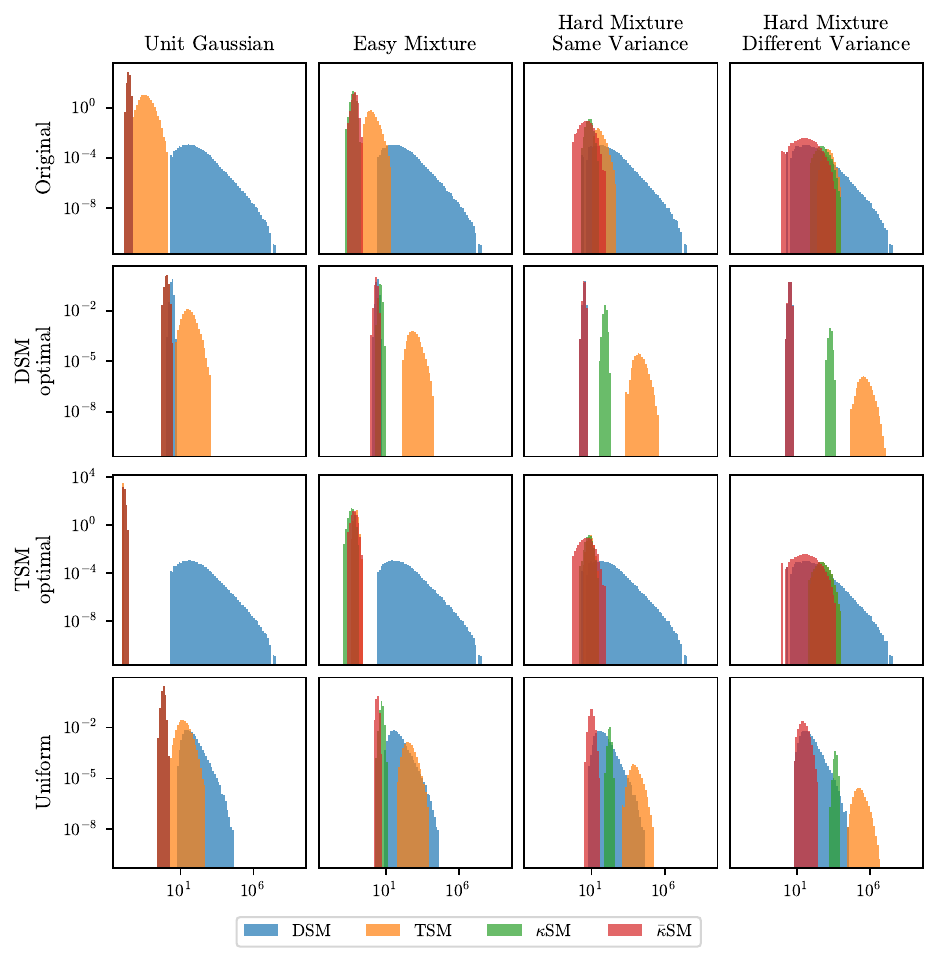}
    \caption{Comparison of the distribution of training losses for the combinations of the 4 target densities, 4 training losses, and 4 weighting functions.}
    \label{fig:sm_distribution}
\end{figure}

Next we investigate the variance in the score matching loss functions derived from the score identities discussed. In this case we need to pick a weighting scheme \(\lambda_t\) across time for the score matching losses.
We investigate four weighting schemes here:
\begin{itemize}
    \item The weighting from \cite{song2020score}, \(\lambda_t = \frac{1}{\sigma_t^2}\).
    \item The weighting from \cite{karras2022elucidating} which ensures for a Gaussian target that the variance of the $\DSM$ loss at initialization is 1, given by \[\lambda_t = \frac{\sigma_t^2}{\sigma_{\text{data}}^2}(\sigma_t^2 + \sigma_{\text{data}}^2).\]We refer to this as the DSM optimal weighting.
    \item A new weighting derived in the spirit of  \cite{karras2022elucidating} which ensures for a Gaussian target that the variance of the $\TSM$ loss is 1 at initialization, given by \[\lambda_t = \frac{\alpha_t^2\sigma_{\text{data}}^2}{\sigma_t^2}(\sigma_t^2 + \alpha_t^2\sigma_{\text{data}}^2).\]We refer to this as the $\TSM$ optimal weighting (see  \Cref{sec:preconditioning_training}).
    \item The uniform weighting, \(\lambda_t = 1\).
\end{itemize}
In addition, we numerically normalise these schedules so that they integrate to 1. This will not change the properties of the loss function, but will re-scale the overall value of the loss, and is done to better compare between losses. Figure \ref{fig:weightings} depicts the different weighting functions across time, for a \(\sigma_\text{data}^2 = 1\), which we ensure all the densities we employ have. From Figure \ref{fig:weightings}, we observe that the $\DSM$ and $\TSM$ optimal weightings respectively up- and down-weight times where Figure \ref{fig:si_variances} shows that the $\DSI$ and $\TSI$ estimators had larger variances. 
Figure \ref{fig:sm_distribution} is produced by looking at the \(4^3\) combinations of the 4 targets, 4 weighting functions and 4 score matching losses. For each histogram in the chart, we estimate the loss function at the true score setting \(s_\theta(x_t,t) = \nabla \log p_t(x_t)\) 10,000 times, using 100 samples from the combined time integral (sampling \(t\) uniformly) and expectations. The distribution of these losses is plotted. 


\subsection{Trained score models}

Finally, we train our models on a $2$ dimensional Gaussian mixture model to illustrate some of the advantages of our approach. We consider the four different settings of $\DSM$, $\TSM$, $\kappa$ and $\bar\kappa$ as described before. To parameterize $s_\theta$ we consider a MLP with three layers of size 128, sinusoidal time embedding with embedding dimension $128$. We used the ADAM optimizer with learning rate $10^{-4}$.

\begin{figure}[h]
    \centering
    \includegraphics[width=.48\linewidth]{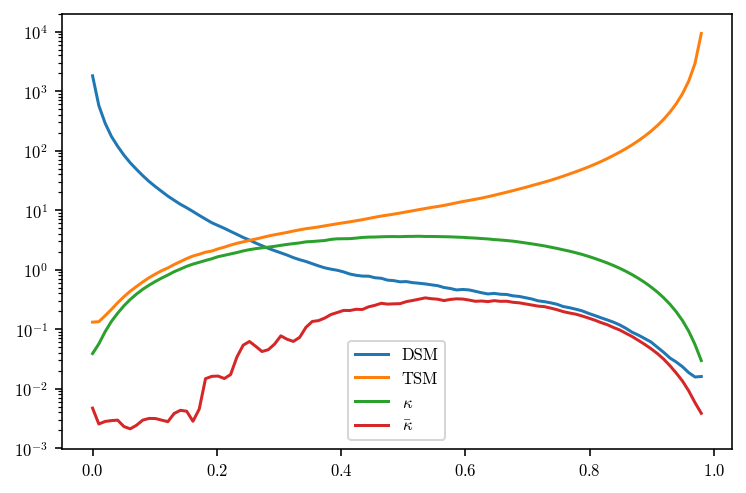} \hfill
    \includegraphics[width=.48\linewidth]{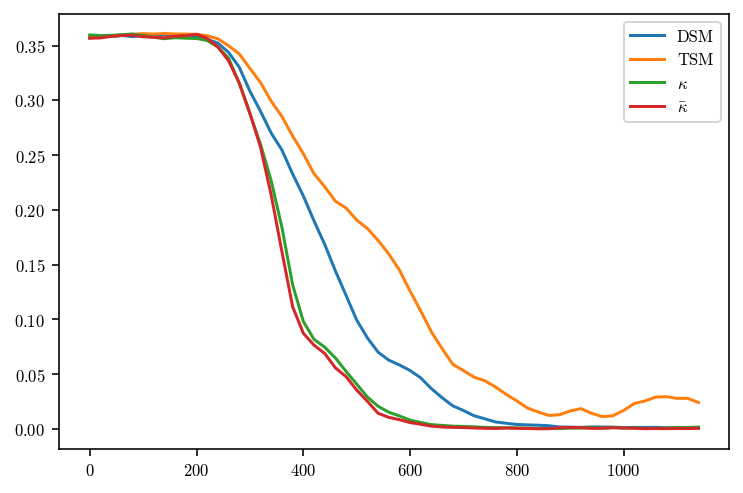}
    \caption{(Left) Mean of the regression loss $\| s_\theta(t, X_t) - L \|^2$ with $\| s_\theta(t, X_t) - L \|^2$ with $L \in \{L_\DSM, L_\TSM, L_{\kappa_t}, L_{\bar\kappa_t} \}$ across training iterations.  (Right) MMD distance between the empirical data distribution and generated samples with score $s_\theta$ for a RBF kernel.}
    \label{fig:training_model}
\end{figure}

In Figure \ref{fig:training_model} (left), we let $L \in \{L_\DSM, L_\TSM, L_{\kappa_t}, L_{\bar\kappa_t} \}$ and compute the mean of the regression loss $\| s_\theta(t, X_t) - L \|^2$.
We consider a uniform weighting strategy for all these losses. The exploding behavior of the $\DSM$ loss at time $0$ and the exploding behavior of the $\TSM$ loss at time $1$ is coherent with the theoretical results of Section \ref{sec:limitations} and  Section \ref{sec:target_informed_score_matching}. Note that only the mixture of targets $\kappa$ and $\bar\kappa$ achieve a non-exploding behavior for all times.
In Figure \ref{fig:training_model} (right), we show the  empirical MMD distance between the empirical data distribution and generated samples with score $s_\theta$ for a RBF kernel. We emphasize the faster convergence of the mixture of targets with $\kappa$ and $\bar\kappa$.

\bibliography{main.bbl}
\bibliographystyle{apalike}

\appendix

\section*{Appendix}

In Appendix \ref{sec:Mainproofs}, we present the proofs of all the Propositions presented in the main paper. In Appendix \ref{sec:FP_proof_ti}, we present another proof of $\TSI$ (equation  \eqref{eq:TIscoreidentity}) for additive Gaussian noise based on diffusion techniques. In Appendix \ref{sec:interpolation} we derive the variance of Monte Carlo estimates of the score based on $\DSI$ and $\TSI$ in the Gaussian case. 
In Appendix \ref{sec:preconditioning_training}, we transpose the analysis of \citet{karras2022elucidating} developed to obtain a stable training loss for DDM to the $\TSM$ loss.

\section{Proofs of the Main Results}\label{sec:Mainproofs} 
\subsection{Proof of Proposition \ref{prop:ti_identity}}
For completeness, we present two proofs of this result (without any claim for originality). 
\paragraph{First proof.} We have $Y=X+W$, $X$ and $W$ being independent, so
\begin{equation}
    p_Y(y)=\int p_X(y-w) p_W(w)\rmd w,
\end{equation}
hence
\begin{equation}
    \nabla p_Y(y)=\int \nabla \log p_X(y-w) 
    ~p_X(y-w) p_W(w)\rmd w.
\end{equation}
It follows that
\begin{align}
    \nabla \log p_Y(y)&=\int \nabla \log p_X(y-w) \frac{p_X(y-w) p_W(w)}{p_Y(y)}\rmd w\\
    &=\int \nabla \log p_X(x) \frac{p_X(x) p_W(y-x)}{p_Y(y)}\rmd x \\
    &=\int \nabla \log p_X(x)~p_{X|Y}(x|y)\rmd x.
\end{align}

\paragraph{Second proof.} For this alternative proof, it is essential for clarity to emphasize notationally which variable we differentiate with respect to. This proof starts from $\DSI$ and shows that we can recover $\TSI$.  
We have $p_{Y|X}(y|x)=p_W(y-x)$ so by the chain rule
\begin{equation}\label{eq:gradyandgradx}
\nabla_y \log p_{Y|X}(y|x)=- \nabla_x \log p_{Y|X}(y|x) . 
\end{equation}
Now by Bayes rule, we have
\begin{equation}\label{eq:logBayesrule}
\log p_{Y|X}(y|x)=\log p_{X|Y}(x|y) + \log p_Y(y) - \log p_X(x).     
\end{equation}
Hence, from \eqref{eq:logBayesrule} and \eqref{eq:gradyandgradx}, we obtain directly 
\begin{equation}\label{eq:derivative1}
\nabla_y \log p_{Y|X}(y|x)=\nabla_x \log p_X(x)- \nabla_x \log p_{X|Y}(x|y).     
\end{equation}
Additionally, we have by the divergence theorem
\begin{equation}\label{eq:derivative2}
    \int \nabla_x \log p_{X|Y}(x|y)~p_{X|Y}(x|y) \rmd x =0.
\end{equation}
The identity \eqref{eq:TIscoreidentity} follows directly by combining \eqref{eq:DSMidentity} with \eqref{eq:derivative1} and \eqref{eq:derivative2}.

\subsection{Proof of Proposition \ref{prop:ti_loss}}
We have 
\begin{align}
    \ell_{\textup{DSM}}(\theta)&= \int \| s_Y^\theta(y) - \nabla \log p_{Y|X}(y|x) \|^2 p_{X,Y}(x,y) \rmd x \rmd y \\
    &= \int \| s_Y^\theta(y) \|^2 p_{Y}(y) \rmd y - 2 \int \langle s_Y^\theta(y), \nabla \log p_{Y|X}(y|x) \rangle p_{X,Y}(x,y) \rmd x \rmd y \\
    & \qquad \qquad \qquad + \int \| \nabla \log p_{Y|X}(y|x) \|^2 p_{X,Y}(x,y) \rmd x \rmd y.
 \end{align}
 From \eqref{eq:derivative1} and \eqref{eq:derivative2}, we get
 \begin{equation}
     \int \langle s_Y^\theta(y), \nabla \log p_{Y|X}(y|x) \rangle p_{X,Y}(x,y) \rmd x= \int \langle s_Y^\theta(y), \nabla \log p_{X}(x) \rangle p_{X,Y}(x,y) \rmd x
 \end{equation}
 so it follows that 
 \begin{align} 
    \ell_{\textup{DSM}}(\theta)& = \int \| s_Y^\theta(y) \|^2 p_{Y}(y) \rmd y - 2 \int \langle s_Y^\theta(y),  \nabla \log p_X(x) \rangle p_{X,Y}(x,y) \rmd x \rmd y \\
    & \qquad \qquad \qquad + \int \| \nabla \log p_{Y|X}(y|x) \|^2 p_{X,Y}(x,y) \rmd x \rmd y \\
&= \int \| s_Y^\theta(y) -  \nabla \log p_X(x) \|^2 p_{Y}(y) \rmd y - \int \| \nabla \log p_X(x) \|^2 p_X(x) \rmd x \\
    & \qquad \qquad \qquad + \int \| \nabla \log p_{Y|X}(y|x) \|^2 p_{X,Y}(x,y) \rmd x \rmd y.
\end{align}
The first term on the r.h.s. is equal to $\ell_{\TSM}(\theta)$ so the result follows.
\subsection{Proof of Proposition \ref{prop:generalnoise}}
Combining \eqref{eq:marginalY} and \eqref{eq:generalnoise}, one has
\begin{align}
    p_Y(y) &= \int p_X(x) F(\Phi(y, x)) \rmd x \\
    &=  \int p_X(\Phi^{-1}(y, z)) F(z) |\det(\nabla_2 \Phi(y, \Phi^{-1}(y, z))^{-1})| \rmd z \\
    &= \int p_X(\Phi^{-1}(y, z)) |\det(\nabla_2 \Phi^{-1}(y, z))| F(z)  \rmd z
\end{align}
where we use the change of variables $z=\Phi(y,x)$. Hence, we have by the chain rule and the change of variables $x=\Phi^{-1}(y,z)$
\begin{align}
    \nabla \log p_Y(y) &=  \int [\nabla_1 \Phi^{-1}(y, z)^\top \nabla \log p_X(\Phi^{-1}(y, z)) + \nabla_y \log |\det(\nabla_2 \Phi^{-1}(y, z))|] \\
    &  \quad \quad \times p_X(\Phi^{-1}(y, z)) |\det(\nabla \Phi^{-1}(y, z))| F(z) / p_Y(y) \rmd z \\ 
&=  \int [\nabla_1 \Phi^{-1}(y, \Phi(y, x))^\top \nabla \log p_X(x) \\
& \qquad \qquad  + \nabla_y \log |\det(\nabla_2 \Phi^{-1}(y, \cdot))| (\Phi^{-1}(y, x)) ]  p_{X|Y}(x| y) \rmd x.  
\end{align}

\subsection{Proof of Proposition  \ref{prop:ti_estimator_lie}}
Using that $\mu$ is left-invariant 
\begin{equation}
\label{eq:change_of_variable}
    p_Y(y) = \int p_X(x) F(y^{-1}x) \rmd \mu(x) = \int p_X(R_x(y)) F(x) \rmd \mu(x) ,
\end{equation}
where $R_x(y) = yx$ for any $x, y \in G$. We have that $\rmd R_x(y) \rmd R_{x^{-1}}(yx) = \Id$. 
Therefore for any $y \in G$ and $h \in \mathrm{T}_y(G)$ we have 
\begin{equation}
    \rmd (p_X \circ R_x) (y)(h) = \rmd p_X(yx) \rmd R_x(y)(h) = \langle \nabla p_X(yx) , \rmd R_x(y)(h) \rangle = \langle \rmd R_x(y) \rmd R_{x^{-1}}(yx) \nabla p_X(yx) , \rmd R_x(y)(h) \rangle.
\end{equation}
Finally, using that $\langle \cdot, \cdot \rangle$ is right-invariant we get for any $y \in G$ and $h \in \mathrm{T}_y(G)$ 
\begin{equation}
    \rmd (p_X \circ R_x) (y)(h) = \langle \rmd R_{x^{-1}}(yx) \nabla p_X(yx) , h \rangle .
\end{equation}
Combining this result and \eqref{eq:change_of_variable}  we have 
\begin{align}
    \nabla p_Y(y) &= \int \rmd R_{x^{-1}}(yx) \nabla \log p_X(yx) p_X(yx) F(x) \rmd \mu(x) \\
    &= \int \rmd R_{x^{-1}y}(x) \nabla \log p_X(x) p_X(x) F(y^{-1}x) \rmd \mu(x) . 
\end{align}
Hence, we get that 
\begin{equation}
    \nabla \log p_Y(y) = \int \rmd R_{x^{-1}y}(x) \nabla \log p_X(x) p_{X|Y}(x|y) \rmd \mu(x) . 
\end{equation}

\subsection{Proof of Proposition \ref{prop:interpolant}}

It follows directly from \eqref{eq:interpolant} that
\begin{equation}
    p_Y(y)=\int p_W(y-\alpha x_0-(1-\alpha x_1)) p_{X_0}(x_0)p_{X_1}(x_1) \rmd x_0 \rmd x_1.
\end{equation}
So by considering the change of variables $w=y-\alpha x_0-(1-\alpha)x_1$, i.e. $x_0=\alpha^{-1}(y-(1-\alpha)x_1-w)$, we obtain
\begin{equation}
    p_Y(y)=\alpha^{-1} \int p_{X_0}(\alpha^{-1}(y-(1-\alpha)x_1-w)) p_W(w) p_{X_1}(x_1) \rmd w \rmd x_1.
\end{equation}
so, emphasizing here which variable we differentiate with for clarity, we get 
\begin{align}
    &\nabla_y \log p_Y(y)\\
    =&\alpha^{-1} \int \nabla_y \log p_{X_0}(\alpha^{-1}(y-(1-\alpha)x_1-w)) \frac{p_{X_0}(\alpha^{-1}(y-(1-\alpha)x_1-w)) p_W(w)  p_{X_1}(x_1)}{p_Y(y)} \rmd w \rmd x_1\\
    =&\alpha^{-2} \int \nabla_{x_0} \log p_{X_0}(\alpha^{-1}(y-(1-\alpha)x_1-w))  \frac{p_{X_0}(\alpha^{-1}(y-(1-\alpha)x_1-w)) p_W(w)  p_{X_1}(x_1)}{p_Y(y)} \rmd w \rmd x_1\\
    &=\alpha^{-1} \int \nabla_{x_0} \log p_{X_0}(x_0)  \frac{p_{X_0}(x_0)p_{X_1}(x_1)p_W(y-\alpha x_0-(1-\alpha x_1))}{p_Y(y)} \rmd x_0 \rmd x_1.
\end{align}
The result \eqref{eq:scoreinterpolant1} follows immediately and \eqref{eq:scoreinterpolant2} is obtained similarly.

\section{Fokker--Planck derivation}
\label{sec:FP_proof_ti}
So far our derivation of $\TSI$ \eqref{eq:TIscoreidentity} relies on $\nabla_y \log p_{Y|X}(y|x)=-\nabla_x \log p_{Y|X}(y|x)$. This identity is due to the additive nature of the noising process, i.e. $Y =X + W$ with $W$ independent from $X$. The change of variables used in Proposition \ref{prop:generalnoise} uses implicitly the same property in the case of additive noise. In what follows, we provide another derivation of $\TSI$ leveraging instead the Fokker-Planck equation, the time-reversal of diffusions and backward Kolmogorov equation. We consider the case where $Y = X + W$ with $W \sim \mathcal{N}(0, \sigma^2 I)$.

In our setting, $Y=X_{t_0}$ with $t_0 = \sigma$, $X_0 = X$ and $\rmd X_t = \rmd B_t$, where $(B_t)_{t \geq 0}$ is a $d$-dimensional Brownian motion. For any $t \geq 0$, we denote by $p_t$ the density of $X_t$. The Fokker-Planck equation shows that $(p_t(x))_{t\in[0,t_0]}$ satisfies the heat equation
\begin{equation}
    \partial p_t(x) = \tfrac{1}{2} \Delta p_t(x),\qquad p_0(x)=p_X(x).
\end{equation}
Hence, we have that
\begin{align}
\label{eq:log_evolution}
    \partial \log p_t (x) &= \tfrac{1}{2} \Delta p_t(x) / p_t(x)\\
    &= \tfrac{1}{2} \| \nabla \log p_t(x) \|^2 + \tfrac{1}{2} \Delta \log p_t(x) . 
\end{align}
If we denote $F_t(x) = \nabla \log p_{t_0 -t}(x)$ then we obtain by differentiating \eqref{eq:log_evolution} w.r.t. $x$ that for any $x \in \rset^d$ and $t \in [0,t_0]$
\begin{equation}
    \partial F_t (x) + \nabla F_t(x) F_t(x) + \tfrac{1}{2} \Delta F_t(x) = 0 . 
\end{equation}
Hence, using the backward Kolmogorov equation we get that for any $y \in \rset^d$ 
\begin{equation}
    F_t(y) = \mathbb{E}[F_{t_0}(Z_{t_0}) \ | Z_t = y ] , \qquad \rmd Z_t = F_t(Z_t) \rmd t + \rmd B_t = \nabla \log p_{t_0-t}(Z_t) \rmd t + \rmd B_t .
\end{equation}
As $F_0 = \nabla \log p_Y$ and $F_{t_0} = \nabla \log p_X$, we get that
\begin{equation}
\label{eq:last_step_fp}
    \nabla \log p_Y(y) = F_{0}(y) = \mathbb{E}[F_{t_0}(Z_{t_0}) \ | Z_0 = y ] = \mathbb{E}[\nabla \log p_X(Z_{t_0}) \ | Z_0 = y] .
\end{equation}
Finally, we notice that $(Z_t)_{t \in [0,t_0]} = (X_{t_0 -t})_{t \in [0,t_0]}$ is the time-reversal of $\rmd X_t = \rmd B_t$ \citep{cattiaux2021time}. Hence, we get that $(Z_0, Z_{t_0})$ admits the same distribution as $(Y,X)$. Combining this result and \eqref{eq:last_step_fp} we obtain
\begin{equation}
    \nabla \log p_Y(y) = \mathbb{E}_{X|Y}[\nabla \log p_X(X)],
\end{equation}
which corresponds to $\TSI$.

\section{Combinining $\DSI$ and $\TSI$ Monte Carlo estimates}
\label{sec:interpolation}

We consider a Gaussian target $p_X(x)= \mathcal{N}(x;0, \sigmadata^2 I)$ as well as additive Gaussian noise $p_W(x)=\mathcal{N}(w;0, \sigma^2 I)$. For $\alpha>0$, we set $Y= \alpha X+W  \sim \mathcal{N}(0, (\alpha^2 \sigmadata^2 +\sigma^2) I)$. The posterior density appearing in $\DSI$ and $\TSI$ is given in this case by 
\begin{equation}
p_{X|Y}(x|y)=\mathcal{N}(x;\alpha \sigmadata^2 /(\alpha^2 \sigmadata^2+\sigma^2) y,\sigma^2 \sigmadata^2/(\alpha^2 \sigmadata^2+\sigma^2)I).
\end{equation}
We compute the variance of the Monte Carlo estimates of the score obtaining by averaging $\nabla \log p_{Y|X}(y|X^i)$ ($\DSI$ estimate) and $\nabla \log p_{X}(X^i)/\alpha$  ($\TSI$ estimate) over samples $X^i \sim p_{X|Y}(\cdot|y)$. We have that 
\begin{align}
    \sum_{i=1}^d \Var_{X|Y}(\nabla_{i} \log p_{Y|X}(y|X)) &= \sum_{i=1}^d \Var_{X|Y}((\alpha X-y)/\sigma^2) \\
    &=d \alpha^2 (\sigmadata/\sigma)^2 /(\alpha^2 \sigmadata^2 +\sigma^2).
\end{align}
On the other hand, we have that 
\begin{align}
    \sum_{i=1}^d \Var_{X|Y}(\nabla_{i} \log p_{X}(X) /\alpha ) &= \sum_{i=1}^d (1/\alpha^2) \Var_{X|Y}(-X/\sigmadata^2) \\
    &=d (\sigma/\sigmadata)^2 /(\alpha^2 \sigmadata^2 +\sigma^2).
\end{align}
Hence we have
\begin{equation}\label{eq:smaller_variance}
    \sum_{i=1}^d \Var_{X|Y}(\nabla_{i} \log p_{Y|X}(y|X))\leq  \sum_{i=1}^d \Var_{X|Y}(\nabla_{i} \log p_{X}(X) /\alpha )
\end{equation}
if and only if $\sigmadata^2 \leq \sigma^2/\alpha$. Again, this is aligned with our previous observations. For $\sigma \gg 1$ then the variance of the $\DSI$ estimator is lower than the variance of the $\TSI$ estimator. For $\sigma \ll 1$ then the variance of the $\TSI$ estimator is lower than the variance of the $\DSI$ estimator. 


We now consider any convex combination of the integrands appearing in $\DSI$ and $\TSI$
\begin{align}
    Z &= \kappa \nabla \log p_{Y|X}(Y|X) + (1- \kappa) \nabla \log p_X(X)/\alpha \\
    &= \kappa \alpha X / \sigma^2 - (1-\kappa) X / (\alpha \sigmadata^2) - \kappa Y / \sigma^2 \\
    &= (\kappa \alpha / \sigma^2 - (1-\kappa) / (\alpha \sigmadata^2)) X - \kappa Y / \sigma^2 \\
    &= (1/\alpha \sigmadata^2) (\kappa (1 + \alpha^2 \sigmadata^2 / \sigma^2) - 1) X - \kappa Y / \sigma^2 .
\end{align}
By construction, the expectation of $Z$ under $p_{X|Y}(x|y)$ is equal to the score $\nabla_y \log p_Y(y)$. In order to minimize the variance of $Z$ under  $p_{X|Y}(x|y)$, we set $\kappa = 1 / (1 + \alpha^2\sigmadata^2 / \sigma^2)$. Hence when $\sigma \gg 1$ we get that $\kappa = 1$ and when $\sigma \ll 1$ we get that $\kappa = 0$. In this specific Gaussian setting we have that the estimator has actually zero variance since $Z = - Y / (\alpha^2 \sigmadata^2 + \sigma^2) = \nabla_y \log p_Y(Y)$.

\section{Preconditioning the training loss}
\label{sec:preconditioning_training}

In this section, we follow the analysis \cite[Appendix B.6]{karras2022elucidating} and derive a rescaled training loss for TSM from first principles for the additive model $Y = \alpha X + W$. 

In \cite{karras2022elucidating}, the input of the network is scaled by $c_i$ and the output is scaled by $c_o$. An additional skip-connection is considered with weight $c_s$. Hence we have 
\begin{equation}
s^{\theta}_Y(y)=c_o F_\theta(\sigma, c_i y) + c_s y.
\end{equation}
The total loss is weighted by $\lambda>0$ and we have 
\begin{align}
    \mathcal{L}(\theta) &=\lambda \ell_{\TSM}(\theta)=\lambda \mathbb{E}_{X,Y}[\| c_o F_\theta(\sigma, c_i Y) + c_s Y -\alpha^{-1} \nabla \log p_X(X) \|^2].
\end{align}
In \cite[Appendix B.6]{karras2022elucidating} the hyperparameters $\lambda, c_i, c_o, c_s$ are computed in the case of the DSM loss with $x_0$-prediction using the following principles \begin{enumerate}[label=(\roman*)]
    \item the input of the network $F_\theta$ should have unit variance,
    \item the target of the regression loss should have unit variance,
    \item the effective weighting of the loss defined as $\lambda c_0^2$ should be equal to one,
    \item we choose $c_s$ to minimize $c_o$ so that the errors of the network are not amplified.
\end{enumerate}

For simplicity, we assume that $p_X = \mathcal{N}(0, \sigmadata^2 I)$ and $W \sim \mathcal{N}(0, \sigma^2 I)$. In this case, we have that $\nabla \log p_X(x) = -x / \sigmadata^2$. 
We then obtain
\begin{align}
    \mathcal{L}(\theta) &=  \lambda \mathbb{E}_{X,Y}[\| c_o F_\theta(\sigma, c_i Y) + c_s Y + X / (\alpha \sigmadata^2) \|^2]   \\
    &=  \lambda/(\alpha\sigmadata)^4 \mathbb{E}_{X,Y}[\| - (\alpha \sigmadata)^2 c_o  F_\theta(\sigma, c_i Y) - (\alpha \sigmadata)^2 c_s Y - \alpha X \|^2]   \\
    &=  \lambda' \mathbb{E}_{X,Y}[\| c_o'  F_\theta(\sigma, c_i' Y) + c_s' Y - \alpha X \|^2],
\end{align}
where
\begin{equation}
\label{eq:identification_quantities}
    \lambda' = \lambda/(\alpha \sigmadata)^4 , \quad c_o' = - (\alpha \sigmadata)^2 c_o , \quad c_s' = - (\alpha \sigmadata)^2 c_s , \quad c_i' = c_i . 
\end{equation}
We emphasize that $\lambda' (c_o')^2 = \lambda c_o^2$. Therefore, the rescaled effective weight is the same as the effective weight described in \cite{karras2022elucidating}.

Using \cite[(117), (131), (138),(144)]{karras2022elucidating}, we get 
\begin{align}
    &\lambda' = (\sigma^2 + \alpha^2 \sigmadata^2) / (\alpha \sigma \sigmadata)^2 , \quad c_o' = \alpha \sigma \sigmadata / (\sigma^2 + \alpha^2 \sigmadata^2)^{1/2} , \\
    &c_s' = \alpha^2 \sigmadata^2 / (\alpha^2 \sigmadata^2 + \sigma^2) , \quad c_i = 1 / \sqrt{\sigma^2 + \alpha^2 \sigmadata^2} .
\end{align}
Hence, combining this result and \eqref{eq:identification_quantities} we get that 
\begin{align}
    &\lambda = \alpha^2 \sigmadata^2 (\sigma^2 + \alpha^2 \sigmadata^2) / \sigma^2 , \quad c_o = - \sigma / [\alpha \sigmadata (\sigma^2 + \alpha^2 \sigmadata^2)^{1/2}] , \\
    &c_s = -1 /  (\sigma^2 + \alpha^2 \sigmadata^2) , \quad c_i = 1 / \sqrt{\sigma^2 + \alpha^2 \sigmadata^2} .
\end{align}
Using \cite[(151)]{karras2022elucidating}, we get that the variance of $\lambda \| c_o F_\theta(\sigma, c_i y) + c_s y + x / (\alpha \sigmadata^2) \|^2$ is equal to one for every time $\sigma, \alpha > 0 $, at initialization for $F_\theta = 0$. In the general case, using the hyperparameters $\lambda, c_i, c_s, c_o$ we get 
\begin{align}
    \mathcal{L}(\theta) =  \sigmadata^2 (\sigma^2 + \alpha^2 \sigmadata^2) / \sigma^2 &\mathbb{E}_{X,Y}[\|  \sigma F_\theta(\sigma, Y / (\sigma^2 + \alpha^2 \sigmadata^2)^{1/2}) / [\sigmadata (\sigma^2 + \alpha^2 \sigmadata^2)^{1/2}]  \\ 
    &\qquad + \alpha Y / (\sigma^2 + \alpha^2 \sigmadata^2) +  \nabla \log p_X(X) \|^2]  .
\end{align}
The score network is then given by 
\begin{equation}
    s_Y^\theta(y) = -(1/\alpha)[\sigma F_\theta(\sigma, y / (\sigma^2 + \alpha^2 \sigmadata^2)^{1/2}) / \sigmadata (\sigma^2 + \alpha^2 \sigmadata^2)^{1/2} + \alpha^2 y / (\alpha^2 \sigmadata^2 + \sigma^2)] . 
\end{equation}

\end{document}